\pdfoutput=1

\documentclass[11pt]{article}

\usepackage[preprint]{acl}

\usepackage{times}
\usepackage{latexsym}

\newcommand{\equalcontrib}{\textsuperscript{†}}
\usepackage[T1]{fontenc}

\usepackage[utf8]{inputenc}
\usepackage{microtype}
\usepackage{graphicx}
\usepackage{subfigure}
\usepackage{booktabs} 
\usepackage{inconsolata}
\usepackage{hyperref}
\usepackage{times}
\usepackage{latexsym}
\usepackage{amsmath}
\usepackage{amssymb}
\usepackage{mathtools}
\usepackage{amsthm}
\usepackage{colortbl}
\usepackage{algorithmic}
\usepackage{microtype}

\usepackage[capitalize,noabbrev]{cleveref}

\usepackage{inconsolata}

\usepackage{graphicx}
\usepackage[textsize=tiny]{todonotes}

\title{Improved Supervised Fine-Tuning for Large Language Models to Mitigate Catastrophic Forgetting }

\author{
  \textbf{Fei Ding\equalcontrib},
  \textbf{Baiqiao Wang\equalcontrib}
\\
  \small{
    \textbf{Correspondence:} \href{mailto:email@domain}{dingfei@email.ncu.edu.cn}}
}
 
\begin{document}
\maketitle
\begin{abstract}
Supervised Fine-Tuning (SFT) is a critical step for enhancing the instruction-following capabilities of Large Language Models (LLMs) and adapting them to specialized domains. However, SFT often leads to a degradation of the model's general abilities, a phenomenon known as catastrophic forgetting. This problem is exacerbated when third-party practitioners fine-tune open-source models, as the original SFT data is typically not available. To address this challenge, we propose a novel and cost-effective SFT method that effectively mitigates catastrophic forgetting without requiring access to the original SFT data. Our approach first reconstructs the likely instruction distribution of the base model. It then employs a multi-model generation and filtering pipeline to synthesize a high-quality general-purpose dataset. This synthetic dataset is mixed with new, domain-specific data for fine-tuning. Experimental results show that our method not only preserves the model's capabilities in general domains but also improves task-specific performance, outperforming baselines that use publicly available SFT datasets.
\end{abstract}

\section{Introduction}

Large Language Models (LLMs) have demonstrated remarkable proficiency in understanding and generating human-like text, leading to their widespread adoption across diverse applications. The release of powerful, publicly accessible base models like LLaMA~\citep{touvron2023llama, touvron2023llama2}, Falcon~\citep{penedo2023refinedweb}, and Mistral~\citep{jiang2023mistral} has catalyzed a wave of specialization. To optimize these models for specific domains such as medicine, law, or finance, practitioners often perform Supervised Fine-Tuning (SFT) on domain-specific datasets.

However, this specialization introduces a critical challenge: catastrophic forgetting \citep{MCCLOSKEY1989109}. As the model's parameters adapt to the new data distribution, it tends to lose previously acquired general knowledge, thereby diminishing its performance on a wide range of tasks and limiting its broader applicability.

A common strategy to mitigate catastrophic forgetting is to use rehearsal-based methods, which involve mixing general-purpose data with domain-specific data during fine-tuning. However, the effectiveness of this approach hinges on the quality and distributional similarity of the rehearsal data to the data used for the model's initial alignment. Mainstream open-source models do not disclose their original SFT datasets, forcing practitioners to rely on public SFT datasets. These datasets often diverge significantly from the base model's original SFT data distribution, which can lead to suboptimal performance and even exacerbate knowledge degradation.

To overcome this limitation, we propose a novel, cost-effective rehearsal-based method designed for scenarios where the original SFT data is inaccessible. Our approach comprises two main stages. First, we reconstruct the implicit instruction data distribution of the base model. Second, we leverage this distribution to guide the synthesis of a high-quality, general-purpose SFT dataset using a multi-model generation and filtering process. This synthetic dataset serves as a proxy for the original alignment data and is then mixed with new domain-specific data for fine-tuning. Our method is also compatible with existing continual learning frameworks, offering a robust solution for sequential fine-tuning.

Our contributions are as follows:
\begin{itemize}
    \item We propose a novel method to reconstruct the latent instruction distribution of an aligned LLM by analyzing its generative behavior, enabling the creation of a general-purpose SFT dataset that mitigates catastrophic forgetting.
    \item We introduce a \textbf{Multi-Response Generation and Filter} pipeline that uses a committee of powerful LLMs to generate and score responses, significantly improving the quality and diversity of the synthetic SFT data.
    \item Experimental results demonstrate that our method significantly reduces the decline in an LLM's general capabilities after domain-specific fine-tuning compared to standard approaches that mix in third-party datasets.
\end{itemize}

\section{Related Work}

\subsection{Catastrophic Forgetting in LLMs}
Continual learning aims to enable models to learn from a continuous stream of data without forgetting previously acquired knowledge. A primary obstacle in this field is catastrophic forgetting, the sharp decline in performance on past tasks after a model is trained on new data. As the model updates its parameters to fit the new data, it can overwrite the weights crucial for performing earlier tasks. In the context of LLMs, this manifests as a loss of general knowledge and reasoning skills after fine-tuning on a narrow, specialized dataset.

To combat catastrophic forgetting, researchers have proposed several categories of methods: Replay-Based~\citep{shin2017continual,ren2024analyzing}, Regularization-Based~\citep{mi2020continual}, Gradient-Based~\citep{lee2021sequential}, and Architecture-Based~\citep{geng2021continual} approaches. Our work falls under the replay-based paradigm but innovates by synthesizing the replay data directly from the model to be trained.

\subsection{Data Rehearsal}
Robins \cite{rehearsal} introduced the concept of rehearsal shortly after the identification of catastrophic forgetting. Rehearsal aims to mitigate forgetting by re-exposing the model to data or representations from previously learned tasks. Common rehearsal methods include Experience Replay~\cite{rolnick2019experience}, storing subsets of past task examples and periodically retraining the model on a mixture of these historical examples and new data, and Generative Replay~\cite{shin2017continual}, using generative models to create synthetic examples of prior tasks for integration into new task training. A recent innovation is Self-Synthesized Rehearsal (SSR)~\cite{huang2024mitigating}, which continuously leverages LLM checkpoints from prior stages to synthesize and refine input instances, blending them with current-stage data.

\section{Methods}
Our proposed framework aims to create a high-quality, general-purpose dataset that mimics the base model's original alignment data. This process consists of two main stages: (1) reconstructing the general-purpose SFT data from the base LLM and (2) mixing this reconstructed data with new domain-specific data for fine-tuning. 

\subsection{Reconstructing General-Purpose SFT Data}
The reconstruction process involves three automated steps: instruction generation, multi-response generation, and high-quality response filtering.

\subsubsection{Step 1: Instruction Generation}
The goal of this step is to produce a set of instructions $\{I_1, I_2, \dots, I_N\}$ whose distribution approximates that of the instructions used in the base model's original SFT phase. We hypothesize that an aligned model's generative priors are shaped by its training data. Therefore, by prompting the model to generate instructions, we can sample from a distribution that is implicitly conditioned by its alignment.

Given an open-weight aligned LLM, $M_{\text{base}}$ (e.g., Llama-3-70B-Instruct), we use a simple pre-query template to elicit instruction generation.
Specifically, we feed the model the prompt corresponding to the user's turn in its conversational template, such as 
\texttt{"<|start\_header\_id|>user<|end\_header\_id|>"} for Llama-3.
The model's continuation is treated as a generated instruction.
We repeat this process $N$ times to obtain a large corpus of instructions, 
$D_{\text{instr}} = \{I_1, \dots, I_N\}$. 
In our experiments, we set $N = 100,\!000$ to ensure a diverse and representative sample.

\subsubsection{Step 2: Multi-Response Generation}
Once we have the set of instructions $D_{instr}$, the next step is to generate high-quality responses for them. A single model, even the base model itself, may produce suboptimal or generic responses. To enhance quality and diversity, we employ a multi-model generation strategy.

For each instruction $I_j \in D_{instr}$, we query a committee of $K$ powerful LLMs, $\{M_1, M_2, \dots, M_K\}$, to generate responses. In our setup, we use the base model $M_{base}$ itself, along with two other state-of-the-art models known for their strong performance (e.g., GPT-4 and Qwen2.5-72B-Instruct). For each instruction $I_j$, each model $M_k$ generates $L$ candidate responses $\{R_{j,k,1}, \dots, R_{j,k,L}\}$ using nucleus sampling to encourage diversity. This results in a total of $K \times L$ candidate responses for each instruction. We use $L=3$, yielding nine responses per instruction.

\subsubsection{Step 3: Filtering High-Quality Responses}
With a pool of candidate responses for each instruction, the final step is to select the best one. We leverage the same committee of LLMs, $\{M_1, \dots, M_K\}$, as judges. For each instruction $I_j$, we task each model $M_k$ with evaluating all $K \times L$ candidate responses.

The evaluation is guided by a carefully designed prompt, shown in \Cref{tab:rating_prompt} in the Appendix. This prompt asks the model to score a response on a 5-point scale based on criteria such as helpfulness, relevance, clarity, and adherence to an AI persona. For each candidate response $R_{j,k',l}$, we obtain a set of scores $\{s_{j,k',l,1}, \dots, s_{j,k',l,K}\}$ from the judging models. The final score for the response is the average:
$$ \bar{s}_{j,k',l} = \frac{1}{K} \sum_{k=1}^{K} s_{j,k',l,k} $$
We then select the response with the highest average score as the optimal response, $R_j^*$:
$$ R_j^* = \underset{k' \in \{1,\dots,K\}, l \in \{1,\dots,L\}}{\text{argmax}} \bar{s}_{j,k',l} $$
The final reconstructed general-purpose dataset is composed of the instruction-response pairs: $D_{recon} = \{(I_1, R_1^*), \dots, (I_N, R_N^*)\}$.

\subsection{Data Mixing and Training}
The reconstructed general-purpose dataset, $D_{recon}$, serves as a high-quality proxy for the original SFT data. To adapt the model to a new domain, we mix $D_{recon}$ with the new domain-specific SFT dataset, $D_{domain}$. The final training dataset is $D_{train} = D_{recon} \cup D_{domain}$. The mixing ratio can be tuned as a hyperparameter, but we found that even a simple concatenation works well. We then perform full fine-tuning of the base model on $D_{train}$.

\section{Experiments}
\subsection{Setup}
Our experiments are designed to evaluate whether fine-tuning with our reconstructed dataset can effectively mitigate catastrophic forgetting in a domain-specialization task. We assess performance on a suite of general-knowledge benchmarks after fine-tuning on medical data.

\textbf{Base LLM.} We use \textbf{Llama-3-70B-Instruct} as our base model. This model has been instruction-tuned by its developers and exhibits strong general capabilities, making it a suitable candidate for studying the effects of catastrophic forgetting.

\textbf{Domain-Specific Data.} We use a proprietary medical question-answering dataset as our target domain data, $D_{domain}$.

\textbf{Baseline Datasets for Rehearsal.} To establish strong baselines, we compare our method against the standard practice of mixing publicly available SFT datasets with the domain-specific data. We selected a diverse set of popular open-source datasets:
\textbf{ShareGPT}~\citep{vicuna2023} and \textbf{WildChat}~\citep{zhao2024wildchat} (human-written);
\textbf{Evol-Instruct}~\citep{xu2023wizardlm}, \textbf{UltraChat}~\citep{ding2023ultrachat}, and \textbf{GenQA}~\citep{chen2024genqa} (synthetic);
\textbf{OpenHermes 1}~\citep{OpenHermes}, \textbf{OpenHermes 2.5}~\citep{OpenHermes2.5}, and \textbf{Tulu V2 Mix}~\citep{tulu2} (mixed-source).
For fair comparison, we follow \citet{meng2024simpo} and use the 208K sanitized version of UltraChat.

\textbf{Evaluation.} We measure the model's performance on a broad range of general-purpose tasks from the Hugging Face Open LLM Leaderboard \citep{open-llm-leaderboard}. The selected benchmarks test different capabilities:
\begin{itemize}
    \item \textbf{MMLU-PRO} \citep{wang2024mmlu}: Advanced reasoning and general knowledge.
    \item \textbf{GPQA} \citep{rein2023gpqa}: Graduate-level and expert-level question answering.
    \item \textbf{IFEval} \citep{zhou2023instruction}: Instruction following capability.
    \item \textbf{MATH (Level 5)} \citep{hendrycks2021measuring}: Complex mathematical reasoning.
\end{itemize}

\textbf{Implementation Details.} For all experiments, we create training datasets of the same total size. The domain-specific medical data constitutes a fixed 17\% of the final mix, with the remaining 83\% being the general-purpose rehearsal data (either from our method or a baseline dataset). We perform full fine-tuning for 1 epoch using the LLaMA-Factory library. The training was conducted on 8 NVIDIA A800-80GB GPUs. We used the AdamW optimizer with a learning rate of $1 \times 10^{-6}$, a global batch size of 64, and a sequence length of 4096.

\begin{table*}[t]
\centering
\caption{Main results comparing our method against the baselines using public SFT datasets for rehearsal. All models were fine-tuned on a mix of general and medical data . Our method achieves the highest average score, demonstrating its effectiveness in mitigating catastrophic forgetting.}
\label{tab:main_results}
\vspace{2mm}
\scalebox{0.95}{
\begin{tabular}{l cccc | c}
\toprule
\textbf{Alignment Setup} & \textbf{MMLU-PRO } & \textbf{GPQA } & \textbf{IFEval } & \textbf{MATH Lvl 5 } & \textbf{Average} \\
\midrule
Llama-3-70B-Instruct (Base Model) & 46.74 & 4.92 & 80.99 & 23.34 & 39.00 \\
\midrule
ShareGPT & 46.14 & 4.31 & 81.31 & 20.24 & 38.00 \\
Evol-Instruct & 45.76 & 4.64 & 82.52 & 22.30 & 38.81 \\
GenQA & 43.33 & 4.48 & 80.43 & 15.41 & 35.91 \\
OpenHermes 1 & 45.31 & 4.21 & 81.91 & 15.52 & 36.74 \\
OpenHermes 2.5 & 45.63 & 4.79 & 82.33 & 15.62 & 37.09 \\
Tulu V2 Mix & 46.47 & 4.19 & 82.69 & 16.62 & 37.49 \\
WildChat & 45.83 & 4.12 & 81.32 & 22.11 & 38.35 \\
UltraChat & 45.15 & 4.08 & 81.57 & 20.31 & 37.78 \\
\midrule
\textbf{Ours} & \textbf{46.73} & \textbf{4.88} & \textbf{81.93} & \textbf{23.29} & \textbf{39.21} \\
\bottomrule
\end{tabular}
}
\end{table*}

\subsection{Results}
The main experimental results are presented in \Cref{tab:main_results}. Our method, which uses the reconstructed dataset for rehearsal, not only prevents a decline in general capabilities but also achieves a slight improvement over the original Llama-3-70B-Instruct model.

As shown in the table, all baseline models that were fine-tuned using public datasets exhibit a drop in average performance compared to the original Llama-3-70B-Instruct model. The degradation is particularly severe for math and reasoning-heavy tasks like MATH and GPQA. For instance, using GenQA results in a drop of over 3 points in the average score. This confirms our hypothesis that fine-tuning with distributionally mismatched data exacerbates catastrophic forgetting.

In stark contrast, the model fine-tuned with our reconstructed dataset achieves an average score of 39.21, surpassing the original model's score of 39.00. Our method nearly perfectly preserves the challenging MMLU-PRO and MATH scores while maintaining strong performance on GPQA and IFEval. This demonstrates the critical importance of using rehearsal data that aligns with the model's original instruction distribution.

\begin{figure}[h]
    \centering
    \includegraphics[width=0.48\textwidth]{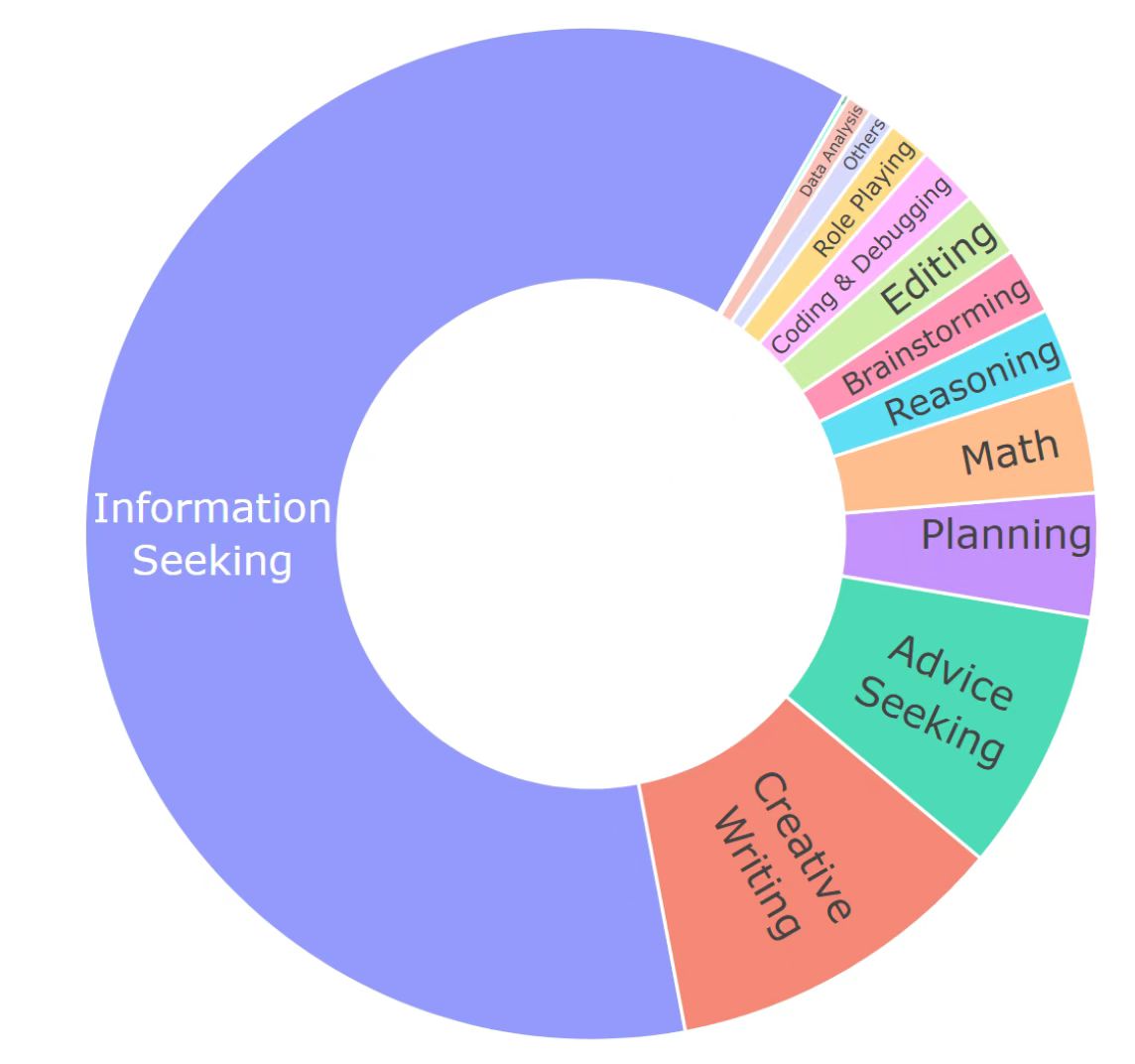}
    \caption{Task category distribution of the SFT data reconstructed from the base LLM (Llama-3-70B-Instruct).}
    \label{fig:distribution}
\end{figure}

\textbf{Impact of SFT Data Distribution.} To understand why our method works, we analyzed the distribution of instruction types in our reconstructed dataset. As illustrated in \Cref{fig:distribution}, the data recovered from Llama-3-70B-Instruct contains a significant proportion of coding, mathematical, and reasoning tasks, alongside more common information-seeking queries. This reflects the diverse and challenging data likely used in its original alignment. In contrast, many public SFT datasets over-represent generic conversational queries and under-represent complex reasoning tasks. This distributional mismatch forces the model's parameters to shift away from the regions that encode these critical skills, leading to performance degradation. Our method succeeds by first recovering a faithful data distribution and then populating it with high-quality content.

\begin{table*}[t]
\centering
\caption{Ablation study results. The full method outperforms the simplified versions, highlighting the benefits of both multi-response generation and using a diverse committee of models for generation and filtering.}
\label{tab:ablation}
\vspace{2mm}
\scalebox{0.85}{
\begin{tabular}{l cccc | c}
\toprule
\textbf{Alignment Setup} & \textbf{MMLU-PRO } & \textbf{GPQA } & \textbf{IFEval} & \textbf{MATH Lvl 5} & \textbf{Average} \\
\midrule
Llama-3-70B-Instruct (Base Model) & 46.74 & 4.92 & 80.99 & 23.34 & 39.00 \\
\midrule
Single-Model-Single-Response & 46.11 & 4.72 & 81.31 & 23.11 & 38.81 \\
Single-Model-Multi-Response-Filtered & 46.55 & 4.91 & 81.72 & 23.06 & 39.06 \\
\midrule
\textbf{Ours (Three-Models-Multi-Response-Filtered)} & \textbf{46.73} & \textbf{4.88} & \textbf{81.93} & \textbf{23.29} & \textbf{39.21} \\
\bottomrule
\end{tabular}
}
\end{table*}

\subsection{Ablation Study}
To dissect the contribution of each component in our method, we conducted an ablation study focusing on the multi-response generation and filtering stages. We tested two simplified configurations:
\begin{itemize}
    \item \textbf{Single-Model-Single-Response}: Instructions are generated as before, but responses are generated only by the base model ($M_{base}$), with one response per instruction and no filtering.
    \item \textbf{Single-Model-Multi-Response-Filtered}: Responses are generated only by the base model ($L=3$), which then also acts as the sole judge to filter for the best response.
\end{itemize}

The results are presented in \Cref{tab:ablation}.
The \textbf{Single-Model-Single-Response} setup, while being the simplest, already achieves an average score of 38.81. This is a strong result, outperforming nearly all public dataset baselines in \Cref{tab:main_results} and underscoring the fundamental importance of getting the instruction distribution right.

Moving to the \textbf{Single-Model-Multi-Response-Filtered} setup, we see a notable improvement to an average score of 39.06. This demonstrates the value of generating multiple candidate responses and applying a quality filter, even when using only the base model. This step helps to mitigate instances where the model might produce a low-quality or incomplete initial response.

Finally, our full method, \textbf{Three-Models-Multi-Response-Filtered}, achieves the best performance at 39.21. The improvement from 39.06 to 39.21 highlights the benefit of using a committee of diverse and powerful models. By incorporating stronger models like GPT-4, we are effectively performing a form of knowledge distillation. These superior models provide higher-quality responses, which serve as better training signals, and also act as more discerning judges, further elevating the quality of the final reconstructed dataset. This hierarchical refinement—first matching the distribution, then enhancing quality through multi-model collaboration—is key to our method's success.

\section{Conclusion}
In this paper, we introduced a novel method to mitigate catastrophic forgetting during supervised fine-tuning when the original alignment data is unavailable. Our approach reconstructs the base model's latent instruction distribution and then synthesizes a high-quality, general-purpose dataset using a multi-model generation and filtering pipeline. When mixed with domain-specific data, this reconstructed dataset allows the LLM to specialize without sacrificing its general capabilities. Our experiments show that this approach is significantly more effective than the common practice of using public SFT datasets for rehearsal. The model fine-tuned with our method not only avoids performance degradation but even surpasses the original base model's average score on a challenging set of general benchmarks.

\section*{Limitations }

We conducted experiments only on Llama-3-70B-Instruct, achieving favorable results. Due to computational constraints, we did not perform extensive testing on other size models and datasets. Furthermore, future work should investigate strategies to seamlessly integrate our approach with existing continual learning methods specifically designed to mitigate catastrophic forgetting. This integration could potentially harness complementary strengths—leveraging our data distribution reconstruction to preserve generalization while employing architectural or regularization-based techniques to stabilize parameter updates—to achieve more robust and comprehensive solutions against knowledge erosion in sequential fine-tuning scenarios.

\section*{Declaration}
Our method is designed to reconstruct an approximate original SFT instruction distribution from the model for enhance SFT performance, not aiming and unable to extract exact training data. Therefore, there is no concern of proprietary data replication.This ensures compliance with ethical and legal standards.

\bibliography{custom}
\newpage

\appendix
\section{Prompt for Response Quality Evaluation}
\label{app:prompt}

The prompt used to have the committee of LLMs score the quality of a generated response is detailed in \Cref{tab:rating_prompt}.

\begin{table*}[h!]
\centering
\caption{The prompt used to evaluate the quality of a response.}
\label{tab:rating_prompt}
\vspace{2mm}
\begin{tabular}{|p{0.9\textwidth}|}
\hline
\begin{minipage}{0.9\textwidth}
\vspace{1em}
Below is a user instruction and an AI response. Evaluate the quality of the AI's response based on how well it fulfills the user's request. Assign a score based on the following 5-point scale:

\textbf{1:} The response is incomplete, off-topic, or contains irrelevant, vague, or missing information. It may repeat the user's question, include personal opinions, or be written from a non-AI perspective (e.g., blog-like). It may also have promotional or irrelevant content.

\textbf{2:} The response addresses some of the user's request but lacks detail or direct relevance. It provides only a general approach instead of a specific solution.

\textbf{3:} The response is helpful but lacks an AI perspective. It covers the user's request but appears taken from a personal blog, webpage, or similar source. It may include personal opinions, experiences, or mentions of external content.

\textbf{4:} The response is clear, complete, and written from an AI's perspective. It directly addresses the user's request, but there may be minor room for improvement, such as clarity or conciseness.

\textbf{5:} The response is excellent, written from an AI's perspective, with a clear focus on the user's request. It is thorough, well-organized, and shows expert knowledge without irrelevant content. The response is logical, easy to follow, and engaging.

Provide a brief justification for your score and then write "Score: \textless rating\textgreater" in the last line.
\\
\\
\texttt{<generated instruction>}
\\
\texttt{<output>}
\vspace{1em}
\end{minipage}
\\
\hline
\end{tabular}
\end{table*}

\end{document}